\pdfoutput=1

\documentclass[11pt]{article}

\usepackage[final]{acl}

\usepackage{times}
\usepackage{latexsym}

\usepackage[T1]{fontenc}

\usepackage[utf8]{inputenc}

\usepackage{microtype}

\usepackage{inconsolata}
\usepackage{amsfonts}
\usepackage{amsmath}
\usepackage{booktabs}
\usepackage{graphicx}
\usepackage{comment}
\usepackage{todonotes}
\usepackage{multirow}
\usepackage{xspace}
\usepackage{xcolor,colortbl}
\usepackage{bm}
\usepackage[normalem]{ulem}
\usepackage{enumitem}
\usepackage{setspace}
\usepackage{url}
\usepackage{supertabular}
\usepackage{array}
\usepackage{mdframed}

\makeatletter
\protected\def\z#1{\pdfsavepos\write\@auxout{\gdef\string\tpos#1{\the\pdflastxpos}}}%
\def\foo#1#2{\ifcsname tpos#1\endcsname\the\dimexpr\csname tpos#2\endcsname sp -\dimexpr\csname tpos#1\endcsname sp\relax\fi}
\makeatother

\newcommand{\eg}{\emph{e.g.,}\ }
\newcommand{\ie}{\emph{i.e.,}\ }
\newcommand{\etc}{\emph{etc.}}
\newcolumntype{x}[1]{>{\centering\arraybackslash\hspace{0pt}}p{#1}}

\title{Internal and External Impacts of Natural Language Processing Papers}

\author{
Yu Zhang \\
Texas A\&M University \\
{\tt yuzhang@tamu.edu}
}

\begin{document}

\maketitle
\begin{spacing}{0.98}

\begin{abstract}
We investigate the impacts of NLP research published in top-tier conferences (\ie ACL, EMNLP, and NAACL) from 1979 to 2024. By analyzing citations from research articles and external sources such as patents, media, and policy documents, we examine how different NLP topics are consumed both within the academic community and by the broader public. Our findings reveal that language modeling has the widest internal and external influence, while linguistic foundations have lower impacts. We also observe that internal and external impacts generally align, but topics like ethics, bias, and fairness show significant attention in policy documents with much fewer academic citations. Additionally, external domains exhibit distinct preferences, with patents focusing on practical NLP applications and media and policy documents engaging more with the societal implications of NLP models.
\end{abstract}

\section{Introduction}
``\textit{Is ACL an AI conference?}''
Comments from opinion leaders\footnote{\url{https://faculty.washington.edu/ebender/papers/ACL_2024_Presidential_Address.pdf}}\footnote{\url{https://gist.github.com/yoavg/f952b7a6cafd2024f44c8bc444a64315}} within the natural language processing (NLP) community definitely matter.
Meanwhile, a complementary perspective is how the broader academic community and the public perceive and utilize papers from top-tier NLP conferences (\eg ACL, EMNLP, and NAACL).
We can categorize the public use of NLP papers into two types:
(1) \textit{Internal Use}: NLP papers are cited by other research articles in not only NLP but also various other fields \cite{zhang2024comprehensive}, due to the ideas, methods, datasets, or findings they present \cite{jurgens2018measuring,cohan2019structural}; and
(2) \textit{External Use}: Research can break out of the academic ``ivory tower'' \cite{cao2023breaking} and diffuse into the technological, societal, and governmental discussions \cite{yin2022public}. Specifically, NLP papers can be referenced by patents, media posts, policy documents, and text in other public channels.
If we infer how public audiences define these NLP conferences based on which papers from the conferences are used by them, the question becomes: Papers of which topics (\eg cutting-edge AI models or linguistic foundations) are more extensively consumed internally and externally?

To answer this question, in this paper, we present a comprehensive scientometric study examining the internal and external impacts of NLP papers published in ACL, EMNLP, and NAACL between 1979 and 2024. We collect data from various sources \cite{priem2022openalex,marx2022reliance,adie2013altmetric,szomszor2022overton} to obtain up-to-date references to NLP papers from internal (\ie research articles) and external (\ie patents, news/social media, and policy documents) domains. Based on the collected data, we calculate the average number of times that papers of a specific topic (\eg ``\textit{Language Modeling}'' or ``\textit{Ethics, Bias, and Fairness}'') are used in an internal/external domain, and employ it as an indicator to measure the impact of the topic within that domain.

Our analysis leads to the following key observations: First, among all research topics that NLP conferences are calling for, language modeling demonstrates the broadest internal and external impacts, whereas linguistic foundations, such as phonology, morphology, psycholinguistics, and pragmatics, exhibit relatively low influence. Second, the internal and external impacts of an NLP topic generally align, though there are outliers, such as ``\textit{Ethics, Bias, and Fairness}'', which show notably high use in policy documents but are cited far less frequently in research articles. Third, different external domains favor different NLP topics, offering complementary insights into an NLP paper's internal impact. Patents tend to consume practical NLP applications, while media and policy documents are more likely attracted by the behavior and societal influence of NLP models. 

\section{Data}
We first describe how we collect data from multiple sources to measure the internal impact of NLP papers and their external uses across various domains, including patents, media, and policy documents.

\vspace{1mm}

\noindent \textbf{NLP Papers: ACL Anthology.} From the ACL Anthology\footnote{\url{https://github.com/acl-org/acl-anthology}, accessed on November 25, 2024.}, we extract all papers whose venue is marked as ACL, EMNLP, or NAACL (including main conference papers, findings papers, system demonstrations, industry track papers, \etc).
We choose these three conferences because CSRankings designates them as ``top conferences'' for ranking purposes, and they are the top three in Google Scholar’s ``Computational Linguistics'' category.
In total, we obtain 24,821 papers published between 1979 and 2024 in these three conferences.

\vspace{1mm}

\noindent \textbf{Internal Impact (Citation): OpenAlex.} Because Microsoft Academic \cite{sinha2015overview} terminated its service at the end of 2021, to get the up-to-date number of citations for each paper, we link our extracted NLP papers to OpenAlex \cite{priem2022openalex}\footnote{\url{https://openalex.s3.amazonaws.com/RELEASE_NOTES.txt}, the version released on November 25, 2024.}. In all, we successfully map 21,104 papers to the OpenAlex database and get their numbers of citations. We use $\mathcal{P}$ to denote the set of these mapped papers. We should mention that each paper $p \in \mathcal{P}$ may have multiple versions (\eg a preprint version and an NLP conference version). When calculating the internal or external impact of $p$, we merge all versions into a single record by summing their citation counts.

\vspace{1mm}

\noindent \textbf{External Impact (Patent): Reliance on Science.} To study references to NLP papers in patents, we follow previous studies \cite{yin2022public,cao2023breaking} and utilize the Reliance on Science dataset \cite{marx2020reliance,marx2022reliance}\footnote{\url{https://zenodo.org/records/11461587}, the version released on June 3, 2024.}, which contains references from USPTO patents to OpenAlex papers. In total, we identify 20,218 links from patents to papers in $\mathcal{P}$.

\vspace{1mm}

\noindent \textbf{External Impact (Media): Altmetric.} To quantify how NLP papers are mentioned in media, we rely on the data provided by Altmetric \cite{adie2013altmetric}\footnote{Please refer to \url{https://www.altmetric.com} to request access to the data. We use the version updated in August 2023.}, which archives references from media, including both news and social media (\ie Twitter, Facebook, Reddit, and blogs), to academic papers.
In all, we obtain 18,586 media-to-paper links for the NLP publications in $\mathcal{P}$.

\vspace{1mm}

\noindent \textbf{External Impact (Policy Document): Overton.} To examine references to NLP papers in policy documents, we follow \citet{yin2021coevolution} and exploit the Overton database \cite{szomszor2022overton}\footnote{Please refer to \url{https://www.overton.io} to request access to the data. We extracted the data on December 8, 2024.}. Overton defines ``policy documents'' as ``research, briefs, reviews, or reports written with the goal of influencing or changing policy'' from ``governments, public bodies, IGOs, NGOs and think tanks''. We obtain 1,223 links from policy documents to papers in $\mathcal{P}$.

\section{Analysis}
\subsection{Internal and External Impacts by Topic}
\label{sec:topic}

\begin{figure*}[t]
\centering
\includegraphics[width=\linewidth]{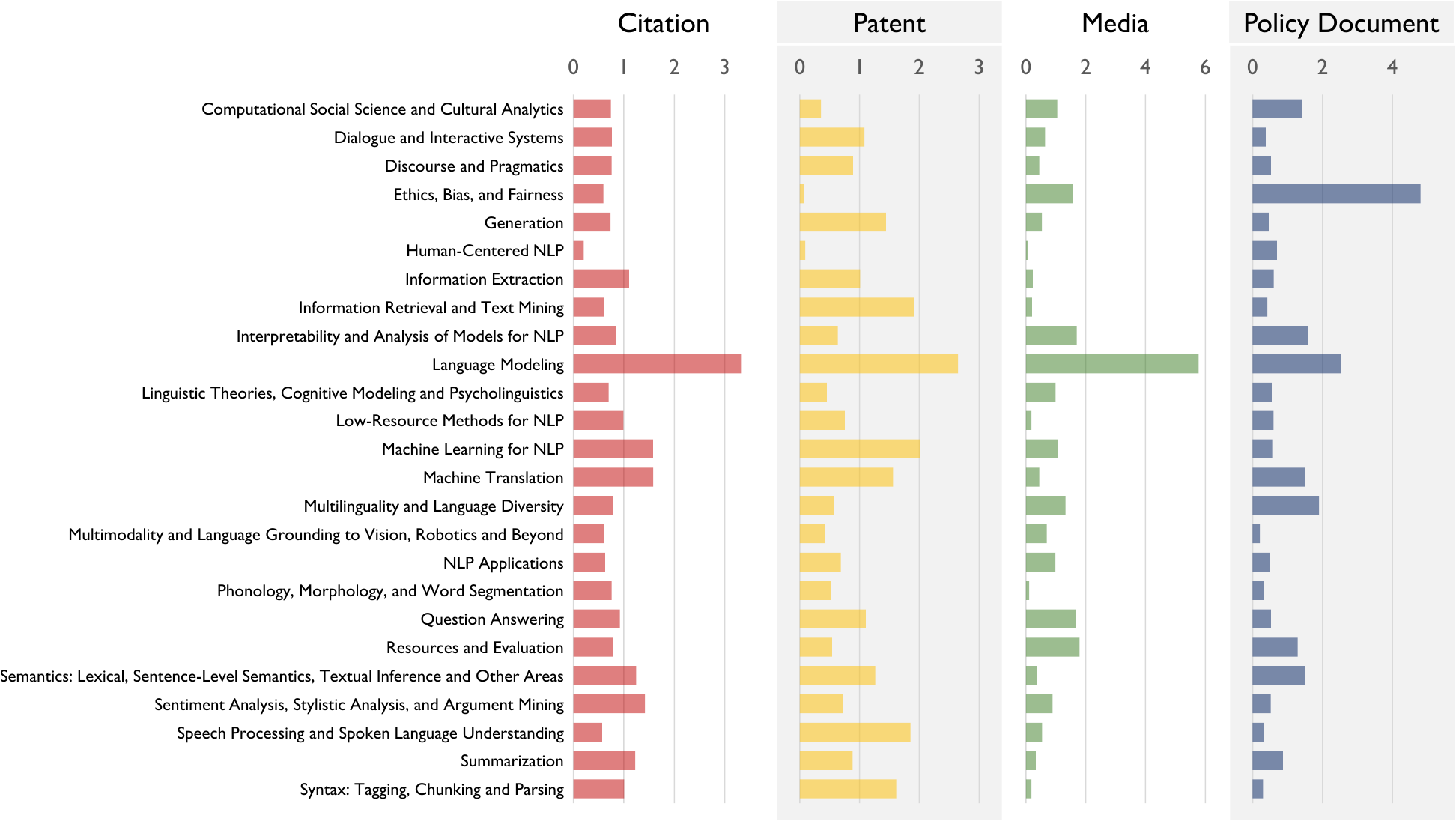}
\caption{Impact of NLP papers with a certain topic in a certain domain.}
\label{fig:bytopic}
\vspace{-0.5em}
\end{figure*}

We adopt the 25 submission topics (excluding the special theme) outlined in the ACL 2025 Call for Papers\footnote{\url{https://2025.aclweb.org/calls/main_conference_papers}} as our topic space for NLP papers, denoted as $\mathcal{T}$. For each NLP paper $p\in \mathcal{P}$, we use GPT-4o \cite{hurst2024gpt} to predict its most relevant topic $t_p \in \mathcal{T}$. Human evaluation is then conducted to assess the quality of GPT-4o annotations. We find that GPT-4o gives reliable annotations, and the human evaluators share ``substantial agreement'' according to the Fleiss' kappa \cite{fleiss1971measuring}. One can refer to Appendix \ref{sec:annotation} for more details about the annotation process. Based on the annotations, we use $\mathcal{P}_t=\{p:p \in \mathcal{P} \textrm{ and } t_p=t\}$ to denote the set of NLP papers labeled with the topic $t$ $(\forall t \in \mathcal{T})$.

To quantify the impact of NLP papers with a certain topic $t \in \mathcal{T}$ in a certain domain $d \in \{\texttt{Citation}, \texttt{Patent}, \texttt{Media}, \texttt{PolicyDocument}\}$, we consider the following metric:
\begin{equation}
{\rm Impact}(t|d) = \frac{\sum_{p \in \mathcal{P}_t} \#{\rm citation}(p|d)/|\mathcal{P}_t|}{\sum_{p \in \mathcal{P}} \#{\rm citation}(p|d)/|\mathcal{P}|}. \notag
\end{equation}
Here, $\#{\rm citation}(p|d)$ represents the ``citation count'' (in a broader sense) of paper $p$ given domain $d$. To be specific, when $d=\texttt{Citation}$, it is the number of times that $p$ is cited by other academic papers (from the entire OpenAlex database rather than ACL Anthology only); when $d \in \{\texttt{Patent}, \texttt{Media}, \texttt{PolicyDocument}\}$, it is the number of times that $p$ is mentioned by patents, media posts, and policy documents, respectively, in our collected data. ${\rm Impact}(t|d)$ is the average ``citation count'' of papers with topic $t$ given domain $d$, normalized by the average ``citation count'' of all papers given domain $d$. Our proposed metric is inspired by \citet{yin2022public}, with one key difference: we consider the actual number of times a paper $p$ is cited, whereas \citet{yin2022public} binarize this value, focusing only on whether $p$ is cited or not (\ie ${\bf 1}_{\#{\rm citation}(p|d) > 0}$).

Figure \ref{fig:bytopic} visualizes ${\rm Impact}(t|d)$ across all topics and domains considered by us, from which we have the following key observations:

\vspace{1mm}
\noindent \textbf{Observation 1}: Papers on language modeling have a broad impact across all internal and external domains.

The topic ``\textit{Language Modeling}'' enjoys the highest impact in the \texttt{Citation}, \texttt{Patent}, and \texttt{Media} domains, as well as the second highest impact in the \texttt{PolicyDocument} domain. It is the only over-represented topic (\ie ${\rm Impact}(t|d) > 1$) across all domains, indicating the significant attention received from a broad range of public audiences on language models.

\vspace{1mm}
\noindent \textbf{Observation 2}: Papers on ethics, bias, and fairness are highly consumed in policy documents but have significantly lower usage in patents and academic papers.

The topic ``\textit{Ethics, Bias, and Fairness}'' achieves the highest impact in \texttt{PolicyDocument}, exceeding the average in \texttt{Media}, while being under-represented in \texttt{Citation} and ranking the last in \texttt{Patent}. The topic ``\textit{Computational Social Science and Cultural Analytics}'' exhibits a similar pattern of external impacts, but with a less pronounced contrast.

\vspace{1mm}
\noindent \textbf{Observation 3}: Linguistic foundations are relatively under-represented in all internal and external domains.

Topics that provide linguistic and theoretical underpinnings of NLP, such as ``\textit{Phonology, Morphology, and Word Segmentation}'', ``\textit{Linguistic Theories, Cognitive Modeling and Psycholinguistics}'', as well as ``\textit{Discourse and Pragmatics}'', have ${\rm Impact}(t|d) < 1$ across all domains.

\vspace{1mm}
\noindent \textbf{Observation 4}: Tasks and techniques benefiting real-world NLP applications have a larger impact in patents.

Besides ``\textit{Language Modeling}'', topics strongly consumed in \texttt{Patent} include ``\textit{Machine Learning for NLP}'', ``\textit{Information Retrieval and Text Mining}'', ``\textit{Speech Processing and Spoken Language Understanding}'', ``\textit{Syntax: Tagging, Chunking and Parsing}'', as well as ``\textit{Machine Translation}''. Many of these topics facilitate practical applications, such as search engines and speech recognition systems.

\vspace{1mm}
\noindent \textbf{Observation 5}: Analyses of the behavior and societal influence of NLP models see stronger adoption in media and policy discussions.

6 topics are over-represented in both \texttt{Media} and \texttt{PolicyDocument}. Apart from ``\textit{Language Modeling}'', the other 5 topics---``\textit{Resources and Evaluation}'', ``\textit{Interpretability and Analysis of Models for NLP}'', ``\textit{Ethics, Bias, and Fairness}'', ``\textit{Multilinguality and Language Diversity}'', as well as ``\textit{Computational Social Science and Cultural Analytics}''---all address the understanding and evaluation of NLP models.

\subsection{Correlation between Internal and External Impacts}
Our next study will draw the following conclusion:

\vspace{1mm}
\noindent \textbf{Observation 6}: NLP papers attracting attention from external domains are more likely to be internally impactful as well.

\begin{table}[t]
\small
\centering
\caption{Pearson correlation coefficient between the impact of NLP papers in the internal domain (\ie \texttt{Citation}) and that in each external domain.}
\resizebox{\linewidth}{!}{
\begin{tabular}{c|ccc}
\toprule
$d$ & \texttt{Patent} & \texttt{Media} & \texttt{PolicyDocument} \\
\midrule
${\rm Corr}({\bf I}_{\texttt{Citation}}, {\bf I}_d)$ & 0.654 & 0.725 & \begin{tabular}[c]{@{}c@{}}0.247\\ (0.599 if excluding \\ ``\textit{Ethics, Bias,} \\ \textit{and Fairness}'') \end{tabular} \\
\bottomrule
\end{tabular}
}
\label{tab:corr}
\vspace{-0.5em}
\end{table}

\vspace{1mm}
In Section \ref{sec:topic}, we have raised a few examples whose internal and external impacts are positively correlated. Now, we quantitatively validate this. 
Intuitively, each domain $d$ can be represented as a vector ${\bf I}_d$, where each entry corresponds to the impact of a specific topic within $d$:
\begin{equation}
    {\bf I}_d = [{\rm Impact}(t|d)]_{t \in \mathcal{T}}. \notag
\end{equation}
Table \ref{tab:corr} presents the Pearson correlation coefficient between ${\bf I}_{\texttt{Citation}}$ and each external ${\bf I}_d$ ($d \in \{\texttt{Patent}, \texttt{Media}, \texttt{PolicyDocument}\}$). We observe strong positive correlation between \texttt{Citation} and \texttt{Patent}/\texttt{Media} (\ie with the Pearson correlation coefficient greater than $0.5$). The correlation between \texttt{Citation} and \texttt{PolicyDocument} is weaker, although still positive. That being said, if we remove ``\textit{Ethics, Bias, and Fairness}'' (which is an evident outlier according to Figure \ref{fig:bytopic}) from consideration, ${\rm Corr}({\bf I}_{\texttt{Citation}}, {\bf I}_{\texttt{PolicyDocument}})$ rises above 0.5 as well.

\begin{table}[t]
\small
\centering
\caption{Hit rate of predicting the top-1\% (internally) highly cited papers when using different external information.}
\resizebox{0.85\linewidth}{!}{
\begin{tabular}{l|c}
\toprule
\textbf{External Domain(s) Considered}  & \textbf{Hit Rate} \\
\midrule
$\emptyset$ & 1.00\%     \\
\midrule
$\{\texttt{Patent}\}$ & 5.46\%     \\
$\{\texttt{Media}\}$ & 9.26\%     \\
$\{\texttt{PolicyDocument}\}$ & 18.29\%    \\
\midrule
$\{\texttt{Patent, Media}\}$ & 26.72\%    \\
$\{\texttt{Patent, PolicyDocument}\}$ & 34.02\%    \\
$\{\texttt{Media, PolicyDocument}\}$ & 45.71\%    \\
\midrule
$\{\texttt{Patent, Media, PolicyDocument}\}$ & 71.88\%    \\
\bottomrule
\end{tabular}
}
\label{tab:hit}
\vspace{-0.5em}
\end{table}

To present more evidence of the positive correlation between internal and external impacts, we follow \citet{yin2022public} and conduct an experiment that leverages external use of NLP research to predict internally most cited papers. Let us consider the top-$1\%$ (internally) highly cited NLP papers in $\mathcal{P}$. If we do not exploit any external signal and randomly pick one paper $p$ from $\mathcal{P}$, the expected probability that $p$ is among the top-$1\%$ cited papers (\textit{a.k.a.,} the hit rate) should be $1\%$. In comparison, Table \ref{tab:hit} shows the hit rate if we consider papers referenced at least once in a specific external domain $d$ (\ie $\#{\rm citation}(p|d) \geq 1$). Papers consumed by patents, media, and policy documents exhibit hit rates of $5.46\%$, $9.26\%$, and $18.29\%$, respectively, which are all large multiples of the $1\%$ baseline.

Combining the results from Tables \ref{tab:corr} and \ref{tab:hit}, we find good alignment between what the public from external domains consume and what is regarded as impactful by researchers themselves.

\subsection{Complementarity of Different External Impacts}
Although we have revealed the positive correlation between \texttt{Citation} and each external domain, the alignment between two external domains does not always hold. Qualitatively, we have discussed the different focal points of \texttt{Patent} and \texttt{Media}/\texttt{PolicyDocument} in Observations 4 and 5 in Section \ref{sec:topic}. Quantitatively, we actually have ${\rm Corr}({\bf I}_{\texttt{Patent}}, {\bf I}_{\texttt{PolicyDocument}}) = -0.140$, implying the complementarity, rather than substitutability, of impacts in patents and policy documents. Table \ref{tab:hit} echos this observation. Indeed, for papers referenced in two external domains, the hit rate increases to $26.72\%$-$45.71\%$, significantly higher than those when we consider a single external domain. A paper referenced in all three external domains is a top-$1\%$ cited paper in $\mathcal{P}$ at an astonishing $71.88$ times the baseline rate. To summarize, we draw the following conclusion:

\vspace{1mm}
\noindent \textbf{Observation 7}: Different external domains may favor different types of NLP papers. Papers attracting attention from multiple external domains are more likely to be internally impactful than those attracting one domain only.

\section{Conclusions}
In this paper, we conduct a large-scale scientometric analysis on the internal and external impacts of NLP papers. We find that the broader academic community and the public's attention to NLP research is primarily driven by the demand for language modeling studies, with the exception of policy-related discussions, which show a greater interest in ethics, bias, and fairness. Although different external domains have varying interests in the specific topics of NLP research, there is an overall positive correlation between the public interests and the impact of papers within the academic ``ivory tower''. By examining how the broader academic community and the public perceive NLP papers, our analysis offers complementary insights into whether ACL can be considered an AI conference.

\section*{Limitations}
This work has the following limitations: First, although the three external domains we consider are all important public spaces, they do not encompass all channels that NLP research may impact. Even within these three domains, NLP papers may be consumed through channels not captured by our collected data. 
Second, we utilize GPT-4o to predict the most relevant topic to each NLP paper. On top of that, we conduct human evaluation for quality assessment. However, there are still chances where GPT-4o's annotations may not be accurate enough, potentially affecting the subsequent analysis. 
Third, we currently lack results (and data) to explain why there is a positive correlation between internal and external impacts. It might be a possible explanation that either researchers or the public are following the other’s preference regarding NLP. \citet{yin2022public} propose another potential hypothesis that each external domain typically involves an ``intermediary'' to engage with science, such as journalists in the media, inventors in the patent field, and policy experts in the government. They leverage their expertise to select scientific results and introduce them into their respective domains.
Finally, since our collected data from certain domains do not indicate when an NLP paper was cited, we are unable to perform a systematic temporal analysis. 
We believe that, with appropriate data support, all of the limitations mentioned above represent highly promising future directions.

\section*{Acknowledgments}
We thank Yian Yin for helpful suggestions.

\bibliography{acl}

\appendix

\setcounter{table}{0}
\renewcommand{\thetable}{A\arabic{table}}

\setcounter{figure}{0}
\renewcommand{\thefigure}{A\arabic{figure}}

\section{Topic Prediction of NLP Papers}
\label{sec:annotation}

\subsection{Topic Annotation using GPT-4o}
We adopt the following instruction prompt to guide GPT-4o in performing topic annotation.

\vspace{2mm}
\begin{mdframed}
\small 
\texttt{Instruction: You will be given the title and abstract of a natural language processing paper, as well as a list of candidate topics. Select the most relevant topic to the given paper from the list.}

\vspace{2mm}
\noindent \texttt{Paper title: \textcolor{gray}{[title; e.g., BERT: Pre-training of Deep Bidirectional Transformers for Language Understanding]}}

\vspace{2mm}
\noindent \texttt{Paper abstract: \textcolor{gray}{[abstract; e.g., We introduce a new language representation model called BERT, which stands for ...]}}

\vspace{2mm}
\noindent \texttt{Candidate labels: ``Computational Social Science and Cultural Analytics'', ``Dialogue and Interactive Systems'', ``Discourse and Pragmatics'', ``Ethics, Bias, and Fairness'', ... \textcolor{gray}{(25 topics)}}
\end{mdframed}

\vspace{2mm}
\noindent We set the temperature as 1.0 (the default setting for GPT-4o in OpenAI’s API). In rare cases, the output of GPT-4o does not exactly match any of the candidate labels. To tackle this, we use a scientific pre-trained language model, SPECTER \cite{cohan2020specter}, to encode the output and each candidate label's name; then, we consider the nearest neighbor of the output in the SPECTER embedding space as the predicted label.

\subsection{Human Evaluation}
To assess the quality of annotations produced by GPT-4o, human evaluation is then conducted. We recruit 3 evaluators, all of whom have rich NLP research experiences. We randomly selected 100 samples for evaluation. For each submission, we ask each annotator to judge the correctness of the label predicted by GPT-4o. The scoring scale is $2$ (the predicted label is the most relevant label to the paper), $1$ (the predicted label is relevant to the paper, although there is a better option), and $0$ (the predicted label is irrelevant to the paper).

According to human evaluation, GPT-4o gets an average score of $1.53$ across the three evaluators. Notably, none of the 100 predictions gets $0$ (\ie irrelevant), indicating the reliability of GPT-4o annotations. We also calculate the Fleiss' kappa \cite{fleiss1971measuring}, which is $0.62$ and implies ``substantial agreement'' among the evaluators.

\section{Impact of NLP Papers in GitHub}
\label{sec:github}

As noted in the Limitations section, the impact of an NLP paper may not be limited to textual citations alone. As a complementary analysis, in this section, we examine the influence of NLP papers in the \texttt{GitHub} domain. Specifically, we analyze the average number of forks for GitHub repositories associated with each NLP topic. It is important to note that we consider this measure of impact to be a mixture of both internal and external influences, as forking a code repository may be done by other researchers for follow-up research, or by practitioners aiming to deploy the work in real-world applications.

\subsection{Data and Metric}
We refer to Papers With Code\footnote{\url{https://production-media.paperswithcode.com/about/links-between-papers-and-code.json.gz}, accessed on March 5, 2025.} to establish the correspondence between our collected NLP papers and their associated GitHub repositories. 7,113 (out of 21,104) NLP papers are mapped to at least one repository in this way. Then, we used the GitHub API\footnote{\url{https://api.github.com/repos}, accessed on March 5, 2025.} to retrieve the number of forks for each corresponding repository. If a paper $p$ is associated with multiple repositories, its fork count $\#{\rm fork}(p)$ is computed as the sum of forks across all associated repositories. Analogous to defining impact through the citation count, we define the impact of NLP papers in \texttt{GitHub} based on the fork count as follows:
\begin{equation}
    {\rm Impact}(t|\texttt{GitHub}) = \frac{\sum_{p \in \mathcal{P}_t} \#{\rm fork}(p)/|\mathcal{P}_t|}{\sum_{p \in \mathcal{P}} \#{\rm fork}(p)/|\mathcal{P}|}. \notag
\end{equation}
Here, we only include NLP papers in $\mathcal{P}$ and $\mathcal{P}_t$ that can be matched to at least one GitHub repository.

\begin{figure}[t]
\centering
\includegraphics[width=\linewidth]{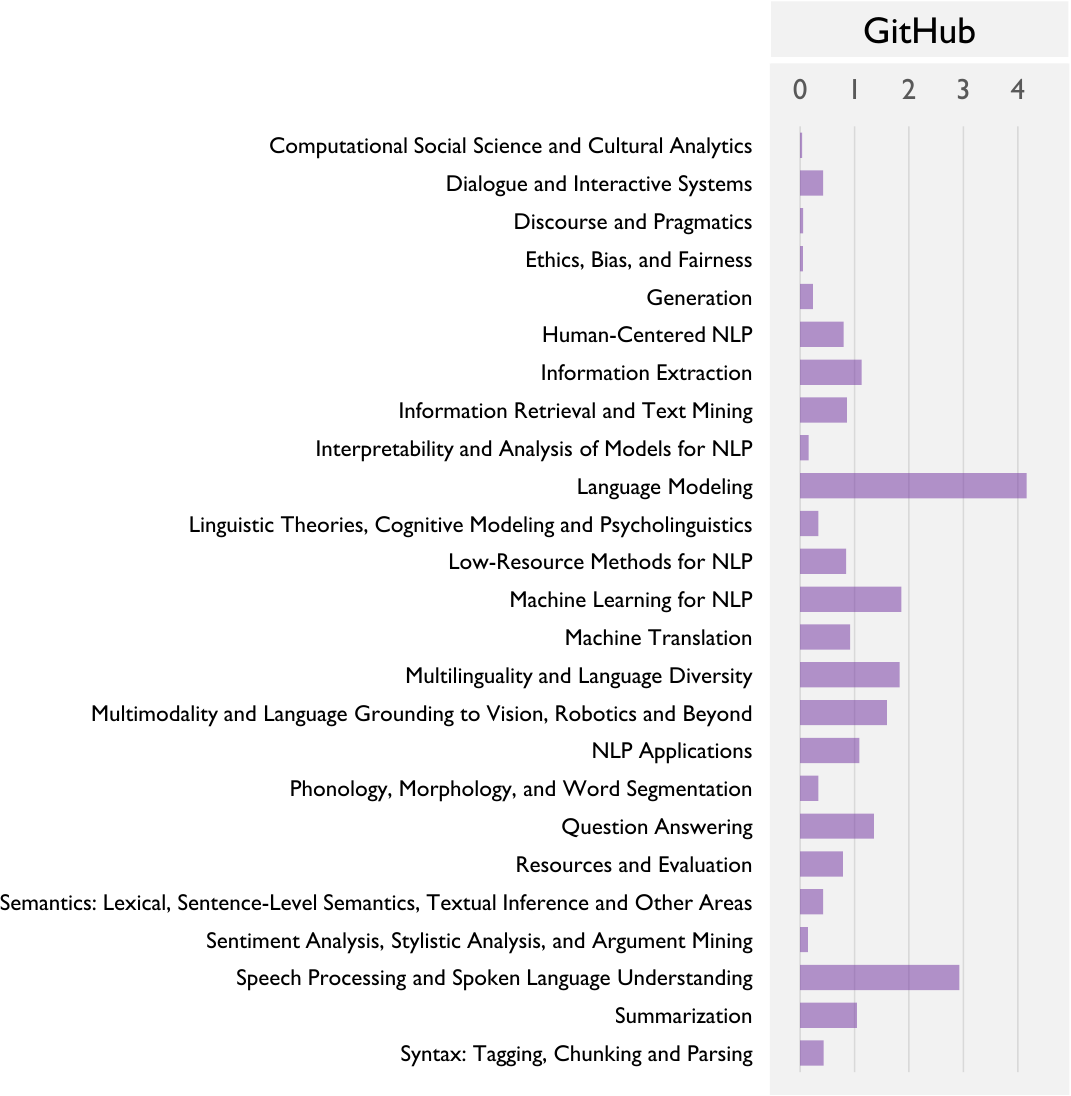}
\caption{Impact of NLP papers with a certain topic in the \texttt{GitHub} domain.}
\label{fig:github}
\end{figure}

\subsection{Analysis}
Figure \ref{fig:github} shows ${\rm Impact}(t|\texttt{GitHub})$ across all topics. Once again, we observe that ``\textit{Language Modeling}'' exhibits the highest impact by a significant margin. Other topics with relatively high impact include ``\textit{Speech Processing and Spoken Language Understanding}'', ``\textit{Machine Learning for NLP}'', ``\textit{Multilinguality and Language Diversity}'', as well as ``\textit{Multimodality and Language Grounding to Vision, Robotics and Beyond}''. It is evident that topics related to practical NLP applications tend to achieve higher impact in \texttt{GitHub}. In this regard, \texttt{GitHub} is  similar to \texttt{Patent}, which also aligns with our intuition. By contrast, linguistic foundations and ``\textit{Ethics, Bias, and Fairness}'' exhibit significantly lower impact.

\begin{table}[t]
\small
\centering
\caption{Pearson correlation coefficient between the impact of NLP papers in \texttt{GitHub} and that in other domains.}
\resizebox{\linewidth}{!}{
\begin{tabular}{c|cccc}
\toprule
$d$ & \texttt{Citation} & \texttt{Patent} & \texttt{Media} & \begin{tabular}[c]{@{}c@{}} \texttt{Policy} \\ \texttt{Document} \end{tabular} \\
\midrule
${\rm Corr}({\bf I}_{\texttt{GitHub}}, {\bf I}_d)$ & 0.586 & 0.633 & 0.531 & 0.009 \\
\bottomrule
\end{tabular}
}
\label{tab:corr_github}
\vspace{-0.5em}
\end{table}

Table \ref{tab:corr_github} demonstrates the Pearson correlation coefficient between the impact of NLP papers in \texttt{GitHub} and that in other domains. We observe a clear positive correlation between \texttt{GitHub} and \texttt{Citation}, \texttt{Patent}, and \texttt{Media}, with the strongest correlation observed with \texttt{Patent}, echoing our qualitative discussion above. Meanwhile, the near-zero correlation between \texttt{GitHub} and \texttt{PolicyDocument} reflects their distinct emphases.

\end{spacing}
\end{document}